\documentclass{article}

    \PassOptionsToPackage{numbers, compress}{natbib}

\usepackage[preprint]{neurips_2024}

\usepackage[utf8]{inputenc} 
\usepackage[T1]{fontenc}    
\usepackage{hyperref}       
\usepackage{url}            
\usepackage{booktabs}       
\usepackage{amsfonts}       
\usepackage{nicefrac}       
\usepackage{microtype}      
\usepackage{xcolor}         
\usepackage{algorithm}
\usepackage{algpseudocode}
\usepackage{times}
\usepackage{epsfig}
\usepackage{epigraph}
\usepackage{graphicx}
\usepackage{amssymb}
\usepackage{tabularx}
\usepackage{multirow}
\usepackage{float}
\usepackage{soul}
\usepackage{pifont}
\usepackage{wrapfig}
\usepackage{threeparttable}
\usepackage{mathtools}
\usepackage{xspace}
\usepackage{enumitem}

\usepackage{amsmath,amsfonts,bm}

\def\eqref#1{equation~\ref{#1}}

\def\1{\bm{1}}

\DeclareMathAlphabet{\mathsfit}{\encodingdefault}{\sfdefault}{m}{sl}
\SetMathAlphabet{\mathsfit}{bold}{\encodingdefault}{\sfdefault}{bx}{n}

\def\gL{{\mathcal{L}}}

\def\sR{{\mathbb{R}}}

\newcommand{\ldist}[0]{\mathcal{L}_{\text{SDS}}}
\newcommand{\x}[0]{\mathbf{x}}

\newcommand{\xt}[0]{{\x}_t}
\newcommand{\xs}[0]{{\x}_s}

\def\name{GaussianDreamerPro\xspace}

\title{GaussianDreamerPro: Text to Manipulable \\3D Gaussians with Highly Enhanced Quality 
}

\author{
Taoran Yi$^{1}$, \; Jiemin Fang$^{2}$\footnotemark[2],\; Zanwei Zhou$^{3}$, \; Junjie Wang$^{2}$,\; Guanjun Wu$^{4}$, \\ \textbf{Lingxi Xie$^{2}$,} \; \textbf{Xiaopeng Zhang$^{2}$,} \; \textbf{Wenyu Liu$^1$\footnotemark[3],} \; \textbf{Xinggang Wang$^1$\footnotemark[2],} \; \textbf{Qi Tian}$^{2}$\ \\
$^1$School of EIC, Huazhong University of Science and Technology \;\;
$^2$Huawei Inc.\\
$^3$MoE Key Lab of Artificial Intelligence, AI Institute, Shanghai Jiao Tong University\\
$^4$School of CS, Huazhong University of Science and Technology\\
\texttt{\small\{taoranyi, guajuwu, liuwy, xgwang\}@hust.edu.cn}\\
\texttt{\small\{jaminfong, is.wangjunjie, 198808xc, zxphistory\}@gmail.com}  \\
\texttt{\small SJTU19zzw@sjtu.edu.cn}  \;\;
\texttt{\small tian.qi1@huawei.com}
}

\begin{document}


\maketitle

{
\renewcommand{\thefootnote}{\fnsymbol{footnote}}
\footnotetext[2]{Project lead.}
\footnotetext[3]{Corresponding author.}
}

\begin{abstract}
Recently, 3D Gaussian splatting (3D-GS) has achieved great success in reconstructing and rendering real-world scenes. To transfer the high rendering quality to generation tasks, a series of research works attempt to generate 3D-Gaussian assets from text. However, the generated assets have not achieved the same quality as those in reconstruction tasks. We observe that Gaussians tend to grow without control as the generation process may cause indeterminacy. Aiming at highly enhancing the generation quality, we propose a novel framework named \name. The main idea is to bind Gaussians to reasonable geometry, which evolves over the whole generation process. Along different stages of our framework, both the geometry and appearance can be enriched progressively. The final output asset is constructed with 3D Gaussians bound to mesh, which shows significantly enhanced details and quality compared with previous methods. Notably, the generated asset can also be seamlessly integrated into downstream manipulation pipelines, \textit{e.g.} animation, composition, and simulation \textit{etc.}, greatly promoting its potential in wide applications. Demos are available at \url{https://taoranyi.com/gaussiandreamerpro/}.

\end{abstract}

\begin{figure*}[thbp]
    \centering
    \includegraphics[width=\linewidth]{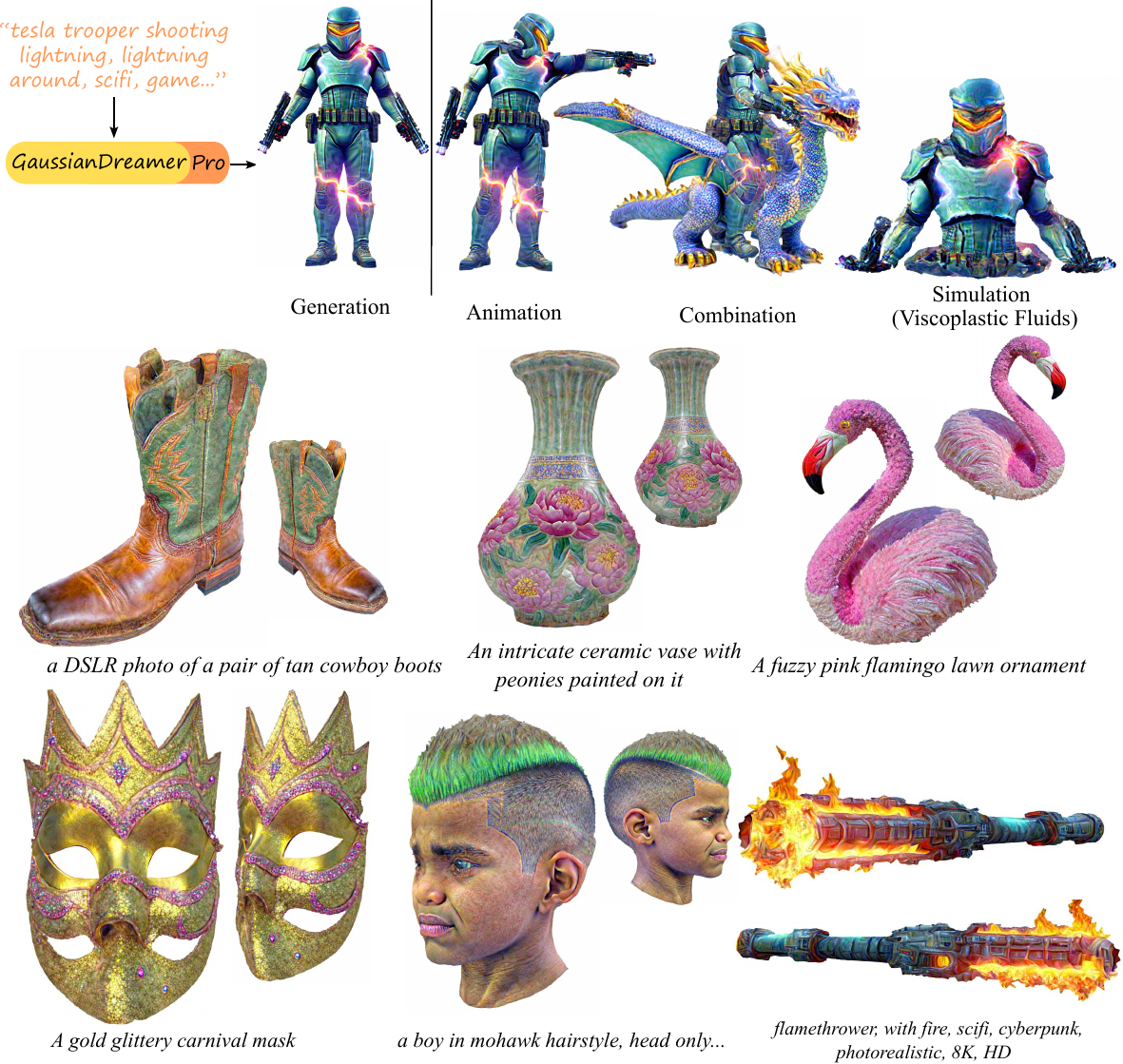}
    \vspace{-20pt}
    \caption{GaussianDreamerPro can generate high-quality 3D assets based on text and can be applied to downstream manipulation pipelines.}
    \label{fig: teaser}
    \vspace{-15pt}
\end{figure*}

\begin{figure}[htbp]
    \centering
    \includegraphics[width=1.0\linewidth]{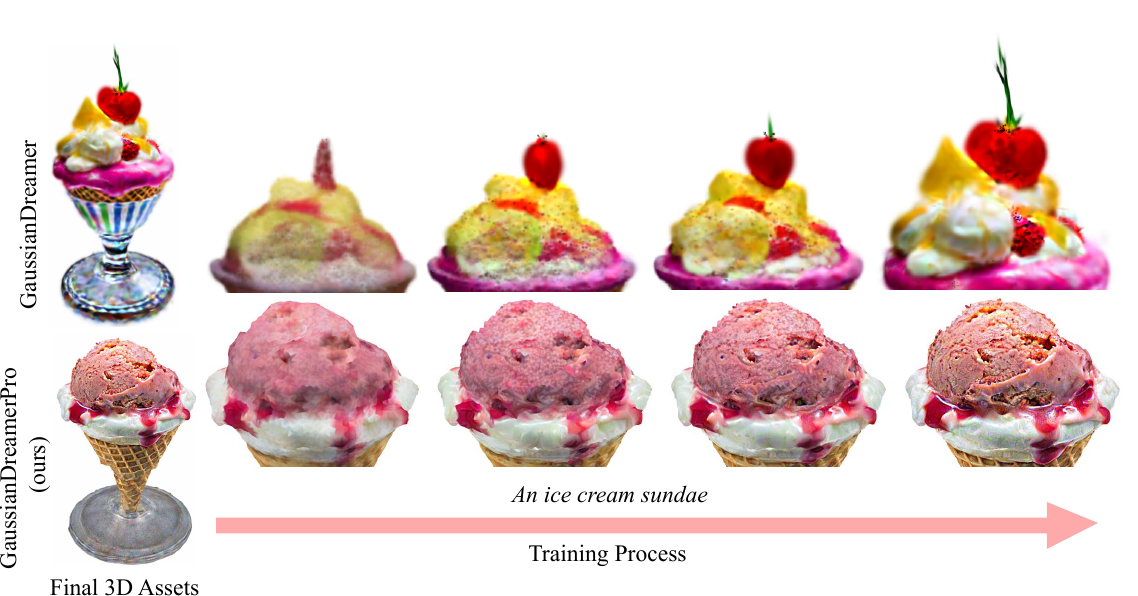}
    \vspace{-15pt}
    \caption{We show the changes in 3D assets during the training process of GaussianDreamer~\cite{yi2023gaussiandreamer} and our method. Compared with GaussianDreamer, which grows Gaussians uncontrollably, resulting in always blurry edges, our method continuously improves the quality of appearance under the constraint of geometry.}
    \label{fig: gs_up}
    \vspace{-10pt}
\end{figure}

\section{Introduction}
The past year has witnessed great success and rapid development of 3D Gaussian splatting (3D-GS)~\cite{kerbl3Dgaussians} in 3D reconstruction and rendering. 3D-GS shows highly realistic rendering effects while enjoying notably fast rendering and training. It has become a popular 3D representation and carries great potential to be applied to various tasks. 

3D generation from text~\cite{yi2023gaussiandreamer,poole2022dreamfusion,chen2023fantasia3d,wang2023prolificdreamer,li2023sweetdreamer,sun2023dreamcraft3d,zhao2023efficientdreamer,shi2023mvdream} is one of the most appealing tasks where researchers have made great efforts to generate high-quality 3D assets, which may benefit the industry of game, movie, and XR \textit{etc}. Many recent works~\cite{yi2023gaussiandreamer,chen2023gsgen,tang2023dreamgaussian,EnVision2023luciddreamer,tang2024lgm,zou2023triplane} expect to transfer the high rendering quality of 3D-GS to the generation task and have made a series of explorations. Though some good 3D assets based on Gaussians can be obtained, the quality still cannot meet the requirements for real applications. E.g., the detail generation is limited and the surfaces of 3D assets are usually blurred as shown in Fig.~\ref{fig: gs_up}. We delve into the problem and analyze the key reasons as follows. Different from the reconstruction task which is based on deterministic information, \textit{e.g.} captured images or videos, the generation task faces more indeterminacy as one text prompt may correlate to various samples with different characters. During the optimization process, 3D Gaussians tend to grow without control towards various directions. This makes it hard to reach a stable state which largely limits the enrichment of details and results in blurred surfaces and appearance.

To tackle the above challenge, we propose a novel text-to-3D generation framework, named \name, aiming at largely enhancing the quality and details of generated 3D-Gaussian assets. The main idea is to bind Gaussians to reasonable geometry which can constrain Gaussians in growing or changing in a controllable range. To make the final generated asset own enough details, we propose to progressively enrich the geometry over 3 stages of generation. Inherited from the basic framework of GaussianDreamer~\cite{yi2023gaussiandreamer}, we first obtain an initial 3D asset with extremely coarse geometry and appearance from the 3D diffusion model, \textit{e.g.} Shap-E~\cite{jun2023shap}. Then we transform the initial asset into a set of 2D Gaussians which can better align the surface and are optimized using the 2D diffusion model. Here we obtain a basic 3D asset whose appearance and geometry are improved to basically meet the requirements. Subsequently, we export a mesh structure with colored vertices from the 2D Gaussians. A series of 3D Gaussians are initialized and bound to the exported mesh. The geometry-bound 3D Gaussians are finally optimized to get the final asset with details highly-enhanced. The asset geometry is progressively evolving through the overall pipeline. Simultaneously, due to the geometry constrain, the appearance can also be enriched continually. Besides, as the final asset is constructed as 3D Gaussians with mesh bound, it can be seamlessly integrated to downstream manipulation pipelines, \textit{e.g.} animation, combination, and simulation \textit{etc}.

Our contributions can be summarized as follows.
\begin{itemize}[leftmargin=10pt]
    \item We design a novel text-to-3D-Gaussians generation framework, which optimize Gaussians with geometry guided, resulting in both the geometry and appearance evolve progressively.
    \item The proposed method can generate 3D assets with quality significantly enhanced compared with previous works.
    \item The generated assets can be seamlessly integrated into downstream manipulation applications. 
    \item The framework is compatible with other 3D generation methods, \textit{e.g.} further enhancing 3D assets generated by DreamCraft3D~\cite{sun2023dreamcraft3d}.
\end{itemize}


\section{Related Work}

\paragraph{Text to 3D Generative Models.}

Recently, text-to-3D generative models ~\cite{hu2023humanliff,gu2023nerfdiff,li20223ddesigner,lin2023magic3d,chen2023fantasia3d,wang2023prolificdreamer,li2023sweetdreamer,sun2023dreamcraft3d,zhao2023efficientdreamer,shi2023mvdream} have made significant progress, achieving high quality in generating both 3D objects and 3D scenes. Some methods~\cite{sanghi2022clip,jain2022zero,michel2022text2mesh,lei2022tango,mohammad2022clip,wang2022clip,xu2023dream3d,hong2022avatarclip} align the 2D images rendered from 3D representations and text with CLIP~\cite{radford2021learning}, thereby generating 3D objects or scenes that match the text description. Currently, there are also some methods~\cite{tang2023dreamgaussian,chen2023gsgen,poole2022dreamfusion,lin2023magic3d,wang2023score,chen2023fantasia3d,wang2023prolificdreamer,xu2023dream3d,shi2023mvdream,zhao2023efficientdreamer,armandpour2023re} that lift 2D diffusion models~\cite{shen2021deep,luo2024measurement,dig} to 3D. Specifically, they bring the distribution of images rendered from 3D representations closer to the distribution of images generated by 2D diffusion models under text conditions through gradient descent, thereby optimizing to obtain 3D representations that meet text conditions. For example, DreamFusion~\cite{poole2022dreamfusion} and \cite{wang2023score} have proposed the Score Distillation Sampling (SDS) and Score Jacobian Chaining (SJC) methods based on this, respectively. This method can achieve good results in generating 3D objects and 3D scenes from text. And because diffusion has a stronger understanding of the text, it can achieve better generation results compared to using CLIP. Some methods~\cite{tang2023make,liu2023one,liu2023zero1to3,qian2023magic123,shi2023zero123++,lin2023consistent123,shi2023toss,sargent2023zeronvs,liu2023syncdreamer,long2023wonder3d,ye2023consistent,weng2023zeroavatar,yang2023consistnet,weng2023consistent123} use 2D diffusion models fine-tuned with camera pose as a condition to achieve 3D generation from a single image. In addition, there are also some methods~\cite{jun2023shap,nichol2022point,gupta3dgen,gao2022get3d,hong2023lrm,li2023instant3d,tang2024lgm,zou2023triplane,shen2024gamba} that generate 3D objects only by inference, by pretraining on 3D data and text pairs, to achieve the effect of directly generating 3D objects from text, which is often much shorter in time than the method of lifting 2D diffusion models to 3D. However, compared to the method of lifting 2D diffusion models to 3D, the training cost of 3D data and text pairs is higher.
\paragraph{3D Representation Methods.}
Recently, Neural Radiance Fields (NeRF)~\cite{mildenhall2020nerf} have achieved great success in representing 3D scenes, achieving realistic and reliable rendering quality. Some variants~\cite{mueller2022instant,chen2022tensorf,sun2022improved,sun2022direct,fridovich2022plenoxels,chen2024survey} use explicit representation to greatly increase the optimization speed, compensating for the optimization speed issue of NeRF. There are also some methods~\cite{barron2021mip,barron2022mip} that further improve the quality of NeRF. Many methods apply NeRF and its variants to the field of 3D generation, achieving high-quality results. In addition, Magic3D~\cite{lin2023magic3d}, Fantasia3D~\cite{chen2023fantasia3d}, etc., introduce the explicit optimizable representation DMTET~\cite{shen2021deep}  into 3D generation, achieving higher quality than using NeRF due to its geometry constraints. Recently, 3D Gaussian Splatting (3D-GS)~\cite{kerbl3Dgaussians}, as a new 3D representation method, has further accelerated the rendering speed and quality. Some variants of 3D Gaussian Splatting~\cite{2dgaussians,guedon2023sugar,duisterhof2023md} improve the geometry and rendering quality. After introducing 3D Gaussians in the field of 3D generation, the 3D generation based on 3D Gaussian Splatting can achieve real-time rendering speed while having high-quality rendering effects. DreamGaussian~\cite{tang2023dreamgaussian} uses 3D Gaussians  as the optimized 3D representation, further exported as a mesh with optimized texture. GaussianDreamer~\cite{yi2023gaussiandreamer} and GSGEN~\cite{chen2023gsgen} use the characteristics of 3D Gaussians to apply 2D and 3D diffusion models together in the 3D generation process. There are methods~\cite{guedon2023sugar,svitov2024haha,shao2024splattingavatar,wen2024gomavatar,xie2023physgaussian,jiang2024vr} that combine mesh and 3D Gaussians as 3D representation. This representation can be combined with traditional graphics pipelines, making it more versatile. One concurrent work~\cite{li2024controllable} uses this representation method to generate 3D assets, taking advantage of its geometry prior.

\begin{figure*}[thbp]
    \centering
    \includegraphics[width=\linewidth]{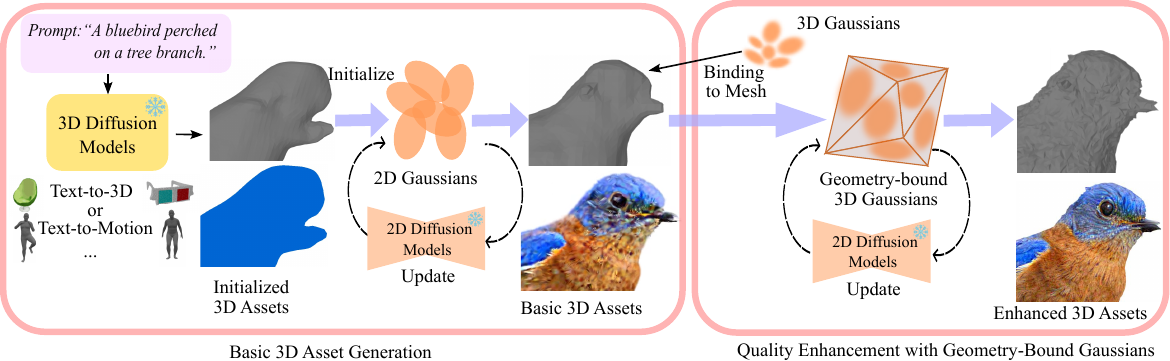}
    \vspace{-12pt}
    \caption{Our framework can be divided into two parts: basic 3D asset generation and quality enhancement with geometry-bound Gaussians. In the basic 3D asset generation stage, we generate initial 3D assets, which are used to initialize 2D Gaussians, obtain basic 3D assets under the optimization of the 2D diffusion model, and export as a mesh. In the quality enhancement with geometry-bound Gaussians stage, we bind 3D Gaussians to the mesh, and also obtain enhanced 3D assets under the optimization of the 2D diffusion model.}
    \label{fig: framework}
\end{figure*}

\section{Method}
\label{Sec:Method}
In this section, we first review the 3D Gaussian splatting~\cite{kerbl3Dgaussians} and 3D generation model. In Sec.~\ref{subsec: overall}, we provide an overview of the whole framework. Then in Sec.~\ref{subsec: basic} and Sec.~\ref{subsec: highly}, we elaborate on the processes of the two stages of generating basic geometry and highly-detailed texturing with 3D Gaussians, respectively.

\subsection{Preliminaries}
\label{Sec:Preliminaries}
\paragraph{3D Gaussian Splatting.}
3D Gaussian Splatting~\cite{kerbl3Dgaussians} is a novel method for representing 3D scenes, achieving excellent rendering quality. It uses a series of 3D Gaussians to represent the scene, each 3D Gaussian contains its center position $\mu_o \in \sR^3$, color $c_o \in \sR^3$, opacity $o_o \in \sR^1$, and covariance $\Sigma_o$. 
For ease of optimization, the covariance matrix is decomposed into a vector $s_o \in \sR^3$ for scaling and a quaternion $q_o \in \sR^4$ to represent rotation. 
Therefore we can represent the 3D Gaussians $\theta_o$ as:
\begin{equation}
    \label{eq: def3dgs}
    \theta_o=(\mu_o, c_o, o_o, \Sigma_o=(s_o, q_o))
\end{equation}
When computing the color $C(r) $ of the corresponding pixel rendered by the ray $r$, 3D Gaussians are splatted into the 2D pixel space. The computation is as follows:
\begin{equation}
	\label{eq: 3dgsrender}
	C(r) = \sum_{i \in \mathcal{N}}c_{i}\alpha_{i}
	\prod_{j=1}^{i-1}(1-\alpha_{j}), \quad
  \alpha_{i} = o_{i} G(x_i), \quad
  G(x_i)~= e^{-\frac{1}{2}x_i^{T}\Sigma^{-1}x_i},
\end{equation}
where $\mathcal{N}$ is the number of 3D Gaussians along the ray $r$ and $c_{i}$ represent the color of the i-th 3D Gaussian, $x_i$ is the offset between the center of the splatted Gaussian and the pixel. $G$ stands for 3D Gaussians' spatial distribution. In 3D Gaussians, the depth of a certain Gaussian in a ray is usually approximated by the depth of its center, which leads to an inaccurate surface.

\paragraph{3D Generation Methods.}
Thanks to the Gaussian Splatting~\cite{kerbl3Dgaussians} representation based on 3D points, GaussianDreamer~\cite{yi2023gaussiandreamer} can be optimized under the geometry guidance provided by the 3D diffusion model. With the geometry guidance provided by the 3D diffusion model, the generated 3D assets have better 3D consistency.
Then the details of 3D Gaussians are further enriched with the 2D diffusion model. Finally, the generated 3D assets have good 3D consistency provided by the 3D diffusion model and fine details from the 2D diffusion model. When enriching the details of 3D Gaussians using the 2D diffusion model $\phi$, GaussianDreamer adopts the SDS loss proposed by DreamFusion~\cite{poole2022dreamfusion}.
In the SDS loss, the 3D representation $\theta$ first renders the image $\x = g(\theta)$ from a given camera pose, where $g$ denotes the differentiable rendering process. Then a random noise $\epsilon$ is added to $\x$ to produce the noisy image $\x_t$. DreamFusion uses a scoring estimation function $\hat\epsilon_\phi(\xt; y, t)$ to predict the sampled noise based on the given noisy image $\x_t$, text condition $y$, and noise level $t$. The difference between the predicted noise $\hat\epsilon_\phi$ and the added random noise $\epsilon$ provides the direction for gradient update:
\begin{align}
    \nabla_{\theta} \ldist(\phi, \x=g(\theta)) \triangleq \mathbb{E}_{t, \epsilon}[w(t)\left(\hat\epsilon_\phi(\xt; y, t)  - \epsilon\right) \frac{\partial\x}{\partial\theta}],
    \label{eq: sdsgrad}
\end{align}
where $w(t)$ is a weighting function. LucidDreamer~\cite{EnVision2023luciddreamer} proposes a multi-step iterative method to predict noise and introduces the (Interval Score Matching) ISM loss, reducing the difficulty of prediction.
LucidDreamer first obtains the predicted noise $\hat\epsilon_\phi(\xs; \emptyset, s)$ at noise level $s = t-\delta_T$, where $\delta_T$ is the Denoising Diffusion Implicit Model (DDIM)~\cite{ddim} inversion step size and $\emptyset$ represents the empty text prompt. Then, through DDIM inversion, $\xt$ is obtained.
Compared to Eq.~\ref{eq: sdsgrad}, the gradient of ISM loss is modified as
\begin{equation}
\label{eq:ISM_grad}
\nabla_{\theta} \gL_{\mbox{\tiny ISM}}(\phi, \x=g(\theta)) \triangleq \mathbb{E}_{t, \epsilon}\,[w(t)(\underbrace{\hat\epsilon_\phi(\x_{t}; y, t) - \hat\epsilon_\phi(\x_s; \emptyset, s)}_{\text{ISM update direction}}) \frac{\partial\x}{\partial\theta}]\mbox{.}
\end{equation}

\subsection{Overall Framework}
\label{subsec: overall}

Our method, as illustrated in Fig.~\ref{fig: framework}, is divided into two stages: \textbf{basic 3D assets genaration} and \textbf{quality enhancement with geometry-bound Gaussians}. In the \textbf{basic 3D assets genaration} stage, We use 2D Gaussians~\cite{2dgaussians} as our 3D representation method, which allows us to obtain a more accurate surface while benefiting from the geometry guidance of the 3D diffusion model. We first use a 3D diffusion model to generate a mesh $M_i$ under a text condition $y$, which serves as the initialization for the 2D Gaussians $\theta_g$. Subsequently, we utilize a text-to-image 2D diffusion model $\phi$ to optimize the 2D Gaussians $\theta_g$ based on the text condition $y$. After the optimization, our 3D representation method exports the 2D Gaussians $\theta_g$ as the basic geometry mesh $M_b$. 
In the \textbf{quality enhancement with geometry-bound Gaussians} stage, we first construct a series of 3D Gaussians $\theta_r$, then bind the 3D Gaussians to the surface of the basic geometry mesh $M_b$. Through the bound 3D Gaussians, we expect high-quality rendering effects while having a more precise geometric structure.
We also leverage the text-to-image 2D diffusion model $\phi$ to optimize the 3D Gaussians, binding the 3D Gaussians $\theta_r$ on the surface of the basic geometry mesh $M_b$ during the optimization process. In the end, we obtain optimized 3D assets based on the 3D Gaussian representation. Throughout this process, from the initialize mesh $M_i $ to the basic geometry mesh $M_b$ exported from the 2D Gaussians $\theta_g$, and finally to the 3D assets represented by 3D Gaussians $\theta_r$, the geometry evolves continuously. Based on this continuously evolving geometry, we achieve 3D assets with progressively improving quality. By binding the 3D Gaussians on the mesh, we can integrate the generated assets into traditional graphics pipelines, making our method applicable to animation and simulation.

\subsection{Basic 3D Asset Generation}
\label{subsec: basic}

3D Gaussian Splatting~\cite{kerbl3Dgaussians} and their variants serve as explicit representation methods that can effectively utilize the coarse geometry generated by 3D diffusion models. However, due to the multi-view inconsistency of 3D Gaussian distributions, 3D Gaussians cannot accurately represent surfaces, resulting in generated 3D assets with rough and imprecise geometry. Inspired by 2D Gaussian Splatting~\cite{2dgaussians}, we flatten the original 3D Gaussians of the ellipse into a 2D surfel to accurately model the geometry of the 3D assets. Different from 3D Gaussians in Eq.~\ref{eq: def3dgs}, the scale vector $s_g$ is represented as $s_g = (s_u, s_v) \in \sR^2$, which controls the variances of the 2D Gaussians. Therefore, the 3D representation to be optimized is denoted as $\theta_g(\mu_g, c_g, o_g, s_g, q_g)$, where $\mu_g$ is the position of the center, $c_g$ is the color, and $o_g$ is the opacity. Its rendering process is similar to Eq.~\ref{eq: 3dgsrender}. 
This representation allows us to leverage the coarse geometry guidance generated by 3D diffusion models while enjoying the precise surface of 3D assets.
Specifically, we first use the 3D diffusion model Shap-E~\cite{jun2023shap} to obtain an initialized mesh $M_i$. Then, we use the mesh $M_i$ to initialize the 2D Gaussians $\theta_g$. The center coordinates $\mu_g$ of 2D Gaussians are obtained from vertices of the mesh $M_i$ vertices as initialization. 
With the geometry guidance of the 3D diffusion model, 2D Gaussians can start from a geometry with good 3D consistency, greatly avoiding the multi-face problem.
Based on the text condition $y$ and given camera poses, we render the image $\x$, randomly give the noise level $s$, get noised images $\x_s$ in Eq.~\ref{eq:ISM_grad}, and get $\x_t$ based on the DDIM inversion process described in Sec.~\ref{Sec:Preliminaries}. Through Eq.~\ref{eq:ISM_grad}, with the text $y$ and the text-to-image diffusion model $\phi$, we can obtain the direction for updating our 3D representation. 
Finally, we perform Poisson reconstruction~\cite{kazhdan2006poisson} based on the center position $\mu_g$ of the 2D Gaussians $\theta_g$, exporting the 2D Gaussians $\theta_g$ as a mesh $M_b$. $v_m$ and $c_m$ denote the vertices of the mesh $M_b$ and the colors of the mesh vertices, respectively. The color $c_m$ is obtained from the color $c_g$ of the Gaussians around the vertices $v_m$. The obtained mesh is represented as follows:
\begin{equation}
	\label{eq: defmesh}
  M_b = (v_m \in \sR^{V \times 3}, c_m \in \sR^{V \times 3}, t_m=(v^1_{m}, v^2_{m}, v^3_{m}) \in \sR^{V \times 3 \times 3}), 
  v_m \leftarrow \mu_g, c_m \leftarrow c_g,
\end{equation}
where $V$ denotes the number of triangles in the mesh $M_b$. $v^1_{m}, v^2_{m}, v^3_{m}$ represent the coordinates of the three vertices that form the triangle $t_m$ of the mesh $M_b$. 
Compared to the elliptical geometry of 3D Gaussians, the surfel geometry of 2D Gaussians greatly reduces the difficulty of calculating level sets, resulting in a better 3D geometry mesh that can be exported.

\subsection{Quality Enhancement with Geometry-Bound Gaussians}
\label{subsec: highly}

\paragraph{Binding Gaussians to Mesh}
In the previous stage, we obtained the basic geometry -- mesh $M_b$. We bind 3D Gaussians on $M_b$ to get high-quality 3D assets. Compared to optimizing texture maps, we hope that optimizing the bound 3D Gaussians can achieve better rendering effects. 
We first construct $V$ clusters of 3D Gaussians $\theta_{r}$, which is the same as the number of triangles in the mesh $M_b$, and each cluster contains $N$ 3D Gaussians. With the help of Eq.~\ref{eq: def3dgs}, we denote the 3D Gaussians as:
\begin{equation}
	\label{eq: defgs}
 \begin{aligned}
 \theta_{r} &= (\theta_{r1}, \theta_{r2}, \dots,\theta_{ri},\dots,\theta_{rV}), \\
  \theta_{ri} &=(\mu_{ri}\in \sR^{N \times 3}, c_{ri}\in \sR^{N \times 3}, \alpha_{ri}\in \sR^{N}, q_{ri}\in \sR^{N \times 4}, s_{ri}\in \sR^{N \times 3}).
  \end{aligned}
\end{equation}
 Here, $i$ represents the i-th cluster of 3D Gaussians $\theta_{r}$. Following Sugar~\cite{guedon2023sugar}, we use frozen barycentric weight $W_b=(W_{b1}, W_{b2}, W_{b3}) \in \sR^{N \times 3}$ and the mesh vertices $v_m$ to calculate the position $\mu_{ri}$ of each Gaussian in the mesh triangle $t_m=(v^1_{m}, v^2_{m}, v^3_{m})$, binding the Gaussian in the triangle. The calculation process is as follows:
\begin{equation}
	\label{eq: posgs}
  \mu_{ri} = W_{b1}* v^1_{m}[i]+ W_{b2}* v^2_{m}[i] + W_{b3}* v^3_{m}[i].
\end{equation}
When the colors $c_{ri}$ of each Gaussian are initializing, it is similar to Eq.~\ref{eq: posgs}, using the colors $c_m$ of the mesh and $W_b$ to calculate.
This ensures that the Gaussians are evenly distributed within the triangle, as can be seen in Fig.~\ref{fig: framework}.


\paragraph{Optimizing Gaussians with Geometry Constraints}
Since $W_b$ here is frozen, 3D Gaussians are tightly bound to the mesh. Rather than directly optimizing the position of each Gaussian, we optimize the coordinates $v_m$ of the mesh vertices. However, the colors $c_{ri}$ of each Gaussian will be directly optimized.
In order to ensure the alignment with the mesh triangle, the rotation of the Gaussian becomes 2D rotation $q_{ri}\in \sR^{N \times 2}$. We maintain the scales $s_{ri}$ in 3D learnable, which ensures that the texture of the mesh surface is more realistic. 
Similar to the previous stage, we also use Eq.~\ref{eq:ISM_grad} to optimize our 3D representation, further improvuing the details of the representation, and obtain 3D assets with 3D Gaussians as textures bound onto the mesh. In the optimization process, our geometry evolves once again, highlighting the details on the surface.


\paragraph{Asset Manipulation}
The position $\mu_{ri}$ of the 3D Gaussian is connected with the mesh vertices $v_m$ using barycentric weights $W_b$. Therefore, when the mesh vertices $v_m$ are changed to $\hat{v_m}$
through the traditional graphics pipeline, such as physical animation, simulation, \textit{etc}.~\cite{jiang2024vr,xie2023physgaussian}, we can calculate the center coordinates $\hat\mu_{ri}$ of the 3D Gaussians $\hat\theta_{ri}$ following the mesh changes through the barycentric weight $W_b$, according to Eq.~\ref{eq: posgs}. 
Finally, through Eq.~\ref{eq: 3dgsrender}, we can get the rendered image of changed 3D Gaussians.

\begin{figure*}[thbp]
    \centering
    \includegraphics[width=0.95\linewidth]{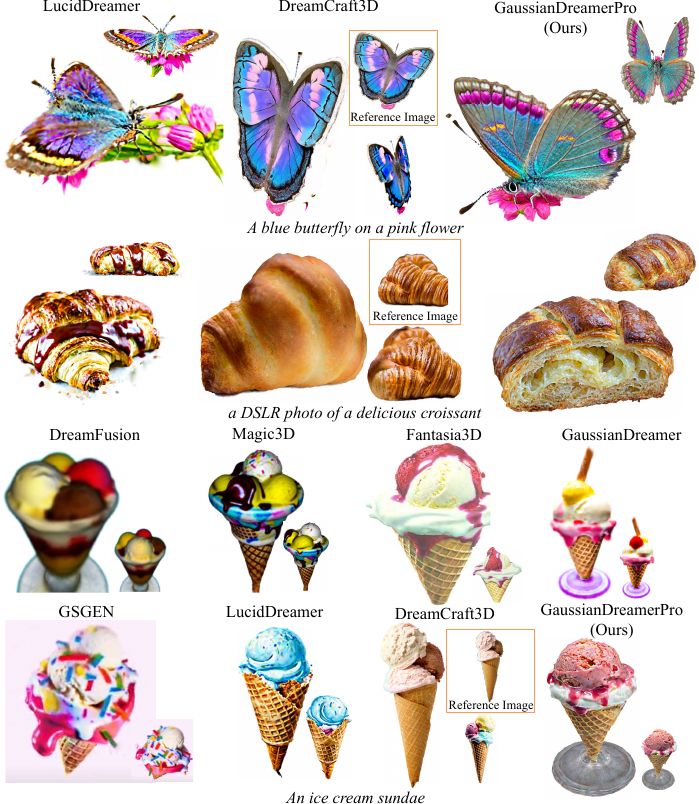}
    \vspace{-5pt}
    \caption{Qualitative comparisons between our method and LucidDreamer~\cite{EnVision2023luciddreamer}, DreamCraft3D~\cite{sun2023dreamcraft3d}, DreamFusion~\cite{poole2022dreamfusion}, Magic3D~\cite{lin2023magic3d}, Fantasia3D~\cite{chen2023fantasia3d} , GaussianDreamer~\cite{yi2023gaussiandreamer} and GSGEN~\cite{chen2023gsgen}.}
    \label{fig: vis}
    \vspace{-25pt}
\end{figure*}

\begin{figure}[t!]
    \centering
    \includegraphics[width=1\linewidth]{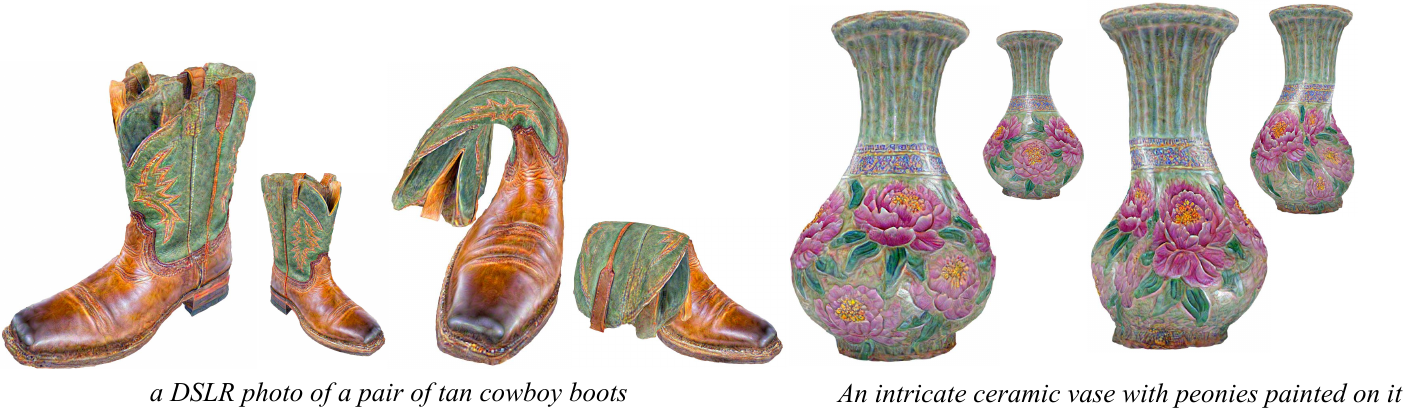}
    \vspace{-16pt}
    \caption{Animation and simulation of the generated 3D assets.}
    \label{fig: ani}
\end{figure}


\section{Experiments}
In this section, we elaborate on the implementation details in Sec.~\ref{subsec: im}. We show the results of a user study in Sec.~\ref{subsec: user}, provide visual results of our method in Sec.~\ref{subsec: vis}, and conduct ablation experiments on relevant parts in Sec.~\ref{subsec: abla}. In the final section, we discuss the limitations of our method.


\subsection{Implementation Details}
\label{subsec: im}

Our method is implemented in PyTorch~\cite{paszke2019pytorch}, using Adam~\cite{kingma2014adam} as the optimizer. The 2D text-to-image diffusion model we use is "stabilityai/stable diffusion-2-1-base"~\footnote{https://huggingface.co/stabilityai/stable-diffusion-2-1-base}~\cite{rombach2022high}, with classifierfree guidance (CFG) size of 7.5. The step size we use is from 0.02 to 0.5. The Shap-E~\cite{jun2023shap} we use loads the model fine-tuned on Objaverse~\cite{deitke2023objaverse} in Cap3D~\cite{luo2023scalable}. The batch size used in the two stages of the framework is 4, the resolution of the rendered image is 1024x1024, and it is downsampled to 512x512 when optimizing with the 2d diffusion model. Both are trained for 5000 iterations. When rendering images, the radius of the camera pose used is from 3.5 to 5.5, azimuth is from -180 degrees to 180 degrees, and elevation is from 30-150 degrees. In the basic geometry generation stage, the learning rate of the position of 2D Gaussians is $1.6 \times 10^{-5}$, and in the quality enhancement with
geometry-bound Gaussians stage, the learning rate of the position of 3D Gaussians is $1.6 \times 10^{-4}$. In addition, the learning rates of color and opacity for both types of Gaussians are $5 \times 10^{-3}$ and $5 \times 10^{-2}$, and the learning rates of scale and rotation are both $5 \times 10^{-4}$. The entire training process is completed on an RTX V100 32G, and it takes about 3 hours.

\subsection{User Study}
\label{subsec: user}

\begin{wrapfigure}{l}{0.5\linewidth}
    \vspace{-12pt}
    \centering
    \resizebox{1.1\linewidth}{!}{
       \includegraphics[width=\linewidth]{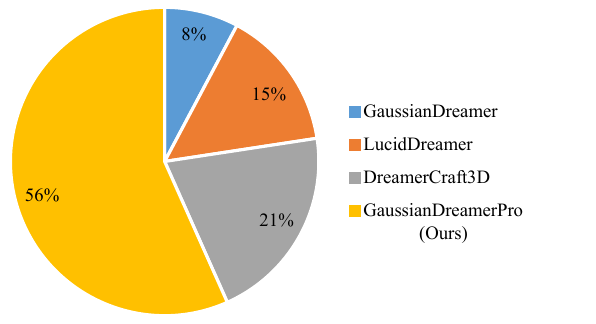}}
    \vspace{-18pt}
    \caption{Results of user study.}
     \label{fig: userstudy}
    \vspace{-5pt}
\end{wrapfigure}
In order to better evaluate the quality of our method, we conducted a series of user studies using 3D assets generated from 10 prompt words. We separately display videos of 3D assets generated by methods GaussianDreamer~\cite{yi2023gaussiandreamer}, LucidDreamer~\cite{EnVision2023luciddreamer}, DreamCraft3D~\cite{sun2023dreamcraft3d}, and our method with the same text prompt. Each participant watches these videos and make a choice on which 3D asset has higher quality. 
We collected 270 responses from 27 participants, where the statistics are organized as Fig.~\ref{fig: userstudy}. Our method wins the favor of 56\% of users, and its performance is much better than other methods. This result indicates that the 3D assets we generate have better quality. 

\subsection{Visualization Results}
\label{subsec: vis}
We present the visualization results of the 3D assets generated by our method, and first compare with the two most recent methods for generating 3D assets -- LucidDreamer~\cite{EnVision2023luciddreamer} and Dreamcraft3D~\cite{sun2023dreamcraft3d}, which are respectively one of the best 3D generative methods based on Gaussian and one of the best 3D generative methods based on optimizable mesh. Dreamcraft3D is different from other methods, it generates 3D assets based on text prompts and a reference image as conditions. Our method is well compatible with the generation method based on optimizable mesh, which we discuss in detail in Sec.~\ref{subsec: abla}. 
Our method shows good quality compared to the other two methods, with higher clarity and better geometry compared to LucidDreamer. In the visualization results of Dreamcraft3D~\cite{sun2023dreamcraft3d}, we have placed the required reference image in the top right, while the other images are the rendering results of the generated 3D assets from Dreamcraft3D. It can be observed that the viewpoint close to the reference image often achieves good results, but the back of the "croissant" lacks detail. Our method has stronger 3D consistency and does not require reference images. Moreover, we provide comparisons with DreamFusion~\cite{poole2022dreamfusion}, Magic3D~\cite{lin2023magic3d}, Fantasia3D~\cite{chen2023fantasia3d}, GaussianDreamer~\cite{yi2023gaussiandreamer}, and GSGEN~\cite{chen2023gsgen}, where DreamFusion uses a NeRF-based~\cite{mildenhall2020nerf,barron2021mip} representation method, Magic3D and Fantasia3D use an optimizable mesh~\cite{shen2021deep} method as the 3D representation method, GaussianDreamer and GSGEN use 3D Gaussians as the representation method, our method shows better geometry and appearance details. Please zoom in Fig.~\ref{fig: vis} for a better comparison. We also show the results of animation and simulation of our generated 3D assets in Fig.~\ref{fig: ani}.

\subsection{Ablation Study and Analysis}
\label{subsec: abla}
\begin{figure}[t!]
    \centering
    \includegraphics[width=\linewidth]{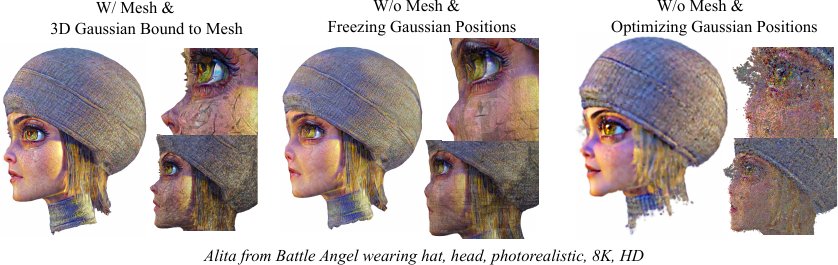}
    \vspace{-10pt}
    \caption{Ablation experiments on geometry constraints.}
    \label{fig: gc}
\end{figure}

In this section, we conduct ablation experiments on the key components of our method to verify the role of key components.
\paragraph{Geometry-Bound Gaussians.}
We show the results of ablating bound geometry in Fig.~\ref{fig: gc}. Each sample's left image is the rendered image, and the right image is the result of visualizing the point cloud saved by the Gaussian coordinates and colors. 
We try to find the connection between the Gaussian position and the rendering quality. In the first one, we bind the Gaussians to the mesh to establish good geometry constraints. In the second one, after initializing the Gaussian positions using Eq.~\ref{eq: posgs}, we removed the mesh. We fix the position of the Gaussians, and the optimization of other parameters is the same as the first one. Compared to the first one, there are fewer geometry constraints here, but freezing the Gaussian positions can to some extent simulate the previous geometry constraints.
Compared to fixing the position of the Gaussians, binding the Gaussians to the mesh surface allows us to optimize the vertices of the mesh to optimize the position of the Gaussians through Eq.~\ref{eq: posgs}. Therefore, we can see that the character's nose and eyes have better quality in the first one. On the last one, we do not add geometry constraints, and directly convert the Gaussians from the basic 3D assets into 3D Gaussians, without the process of geometry-bound, and the other optimization methods are the same. We can see that without geometry constraints, the rendered image is unclear, and we can see that compared to having geometry constraints, there are many discrete Gaussians on the edge and surface of the object, and the position of the Gaussians is very scattered. It is these discrete Gaussians that surround the surface of the object, making the object blurry, and unable to have clear edges and rendering quality. After adding geometry constraints, the growth of Gaussians is controlled, the surface no longer has discrete Gaussians, and clear high-detail images are rendered.

\begin{figure}[t!]
    \centering
    \includegraphics[width=\linewidth]{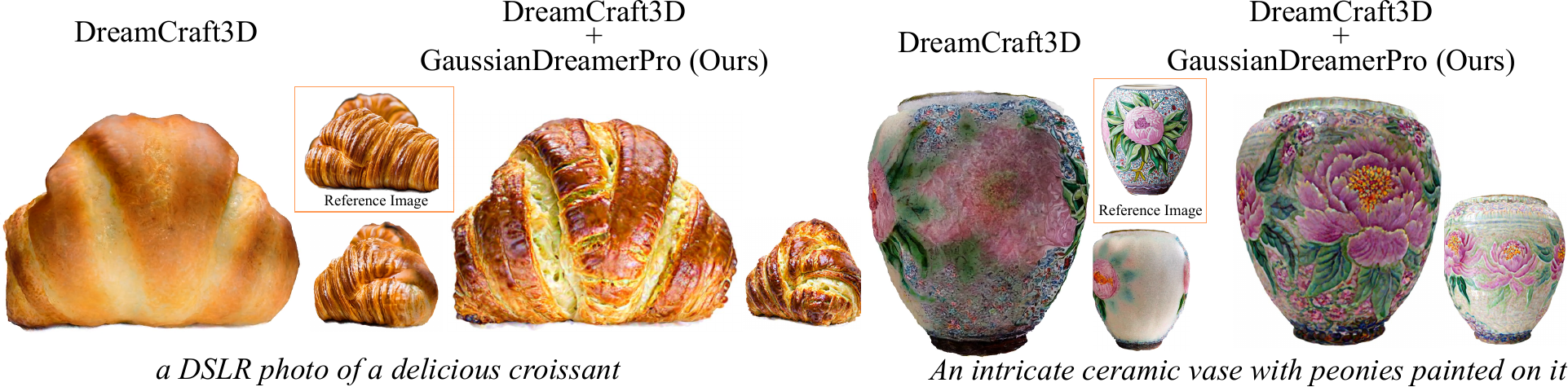}
    \vspace{-10pt}
    \caption{The results of applying our method to DreamCraft3D~\cite{sun2023dreamcraft3d}.}
    \label{fig: dreamcraft3d1}
\end{figure}

\begin{figure}[t!]
    \centering
    \includegraphics[width=\linewidth]{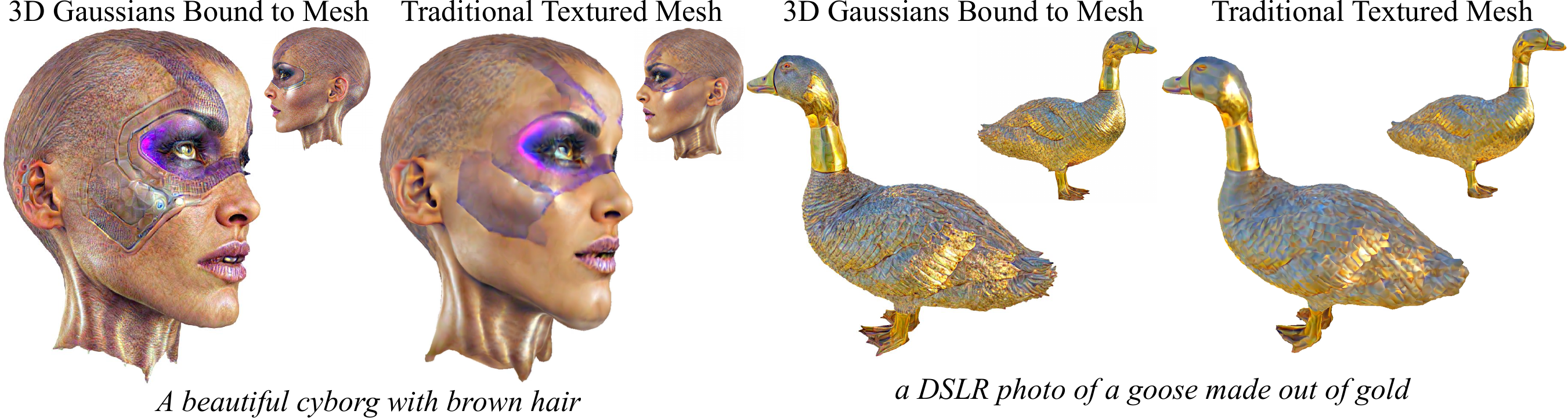}
    \vspace{-12pt}
    \caption{Comparison between enhanced 3D assets optimized with 3D Gaussians bound to mesh and traditional textured mesh.}
    \label{fig: 3dtexture}
\end{figure}

\begin{figure}[t!]
    \centering
    \includegraphics[width=\linewidth]{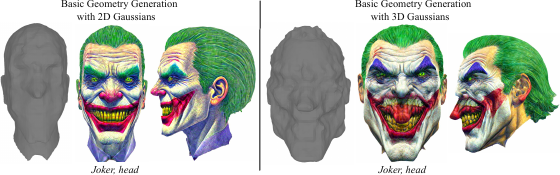}
    \vspace{-12pt}
    \caption{Ablation experiments on the representation method of basic 3D assets.}
    \label{fig: 2dgs}
\end{figure}
\paragraph{Compatibility with Dreamcraft3D}
In this section, we apply GaussianDreamerPro to DreamCraft3D~\cite{sun2023dreamcraft3d}, demonstrating that our method is also friendly to DreamCraft3D. Our results are shown in Fig.~\ref{fig: dreamcraft3d1}. Since the back of the 3D assets generated by DreamCraft3D sometimes suffers from a lack of detail, we use DreamCraft3D as our basic 3D assets and further adopt the method in Sec.~\ref{subsec: highly} for fine-tuning. The results in Fig.~\ref{fig: dreamcraft3d1} show that after fine-tuning, the details on the back of DreamCraft3D can be greatly improved, generating better 3D assets.

\paragraph{Texturing with 3D Gaussians.}

In the quality enhancement with geometry-bound Gaussians stage, we use 3D Gaussians~\cite{kerbl3Dgaussians} as the texture for generating 3D assets. In this section, we conduct ablation experiments to demonstrate that 3D Gaussians as textures have better rendering effects. As comparisons, we change the optimization target of this stage from 3D Gaussians to traditional textured mesh, and we optimize the same number of iterations. Fig.~\ref{fig: 3dtexture} shows our visualization results. Compared to optimizing textured mesh, 3D Gaussians can display higher quality rendering effects. Thanks to the scale of 3D Gaussians in all three dimensions, the 3D assets generated using 3D Gaussians, such as the feathers of the goose in Fig.~\ref{fig: 3dtexture}, have a more realistic texture.

\paragraph{Using 2D Gaussians for Basic 3D Assets Generation.}


In the stage of basic 3D assets generation, we use 2D Gaussians~\cite{2dgaussians} as a 3D representation. In this section, we conduct ablation experiments on this 3D representation, comparing it with the use of 3D Gaussians~\cite{kerbl3Dgaussians}, to verify that it can generate better basic geometry and its impact on the final generation of 3D assets. We present our ablation results in Fig.~
\ref{fig: 2dgs}. Compared to the representation method of using 3D Gaussians in the stage of basic 3D assets generation, using 2D Gaussians can generate better geometry. The visualization quality of the final generated 3D assets is higher, and the consistency is better.

\subsection{Limitations}\label{Sec:limitations}
Our method can generate high-quality single objects based on text input, but when dealing with combinations of multiple objects, it sometimes only generates one of the objects, such as the boots in Fig.~\ref{fig: teaser}. This limitation occurs because the geometry guidance provided by the 3D diffusion model results in the initialized 3D asset. This asset only includes some of the objects described in the prompt when multiple objects are involved. Consequently, these incorrectly initialized 3D assets lead to the incorrectness of final enhanced 3D assets. In future work, it may be possible to achieve better geometry guidance and generate better examples of multiple object combinations by pretraining 3D diffusion models on multiple object dataset.

\section{Conclusion}
In this paper, we propose a novel text-to-3D Gaussian generation framework -- \name, which significantly enhances the quality compared with previous works. The asset geometry is progressively improved over the generation pipeline. With bound to the geometry, 3D Gaussians can be optimized under constraints in a reasonable range, also resulting in continuously enriched appearance. Moreover, the generated assets can be directly applied to various downstream manipulation pipelines. We also demonstrate the universality of the proposed method by further enhancing the quality of assets generated by DreamCraft3D~\cite{sun2023dreamcraft3d}.


{\small
\bibliographystyle{unsrt}
\bibliography{sample-base}
}

\end{document}